%% file: main.tex
\documentclass[10pt,twocolumn,letterpaper]{article}

\usepackage[pagenumbers]{cvpr} 
\usepackage{placeins}
\definecolor{cvprblue}{rgb}{0.21,0.49,0.74}
\usepackage[pagebackref,breaklinks,colorlinks,allcolors=cvprblue]{hyperref}

\usepackage{booktabs}
\usepackage{multirow}
\usepackage{graphicx} 

\newcommand{\shotcell}[2]{%
  \multirow{#1}{*}{\centering\rotatebox[origin=c]{90}{\textbf{#2}}}%
}

\def\confName{CVPR}
\def\confYear{2026}

\title{Learning Multi-Modal Prototypes for Cross-Domain Few-Shot Object Detection}

\author{Wanqi Wang$^{1}$ \quad Jingcai Guo$^2$ \quad Yuxiang Cai$^3$ \quad Zhi Chen$^{4*}$  \\ $^1$University of Chinese Academy of Sciences  \quad $^2$The Hong Kong Polytechnic University  \\  $^3$Zhejiang University  \quad $^4$ The University of Southern Queensland\\
wangwanqi21@mails.ucas.ac.cn, uqzhichen@gmail.com \\ \textbf{* Corresponding Author: Zhi Chen} }

\begin{document}
\maketitle
\input{sec/0_abstract}    
\input{sec/1_intro}
\input{sec/2_formatting}
\input{sec/3_finalcopy}
\input{sec/4_Experiments}

\input{sec/5_Conclusion}
{
    \small
    \bibliographystyle{ieeenat_fullname}
    \bibliography{main}
}
\input{sec/X_suppl}


\end{document}

%% file: sec/0_abstract.tex
\begin{abstract}
Cross-Domain Few-Shot Object Detection (CD-FSOD) aims to detect novel classes in unseen target domains given only a few labeled examples. While open-vocabulary detectors built on vision-language models (VLMs) transfer well, they depend almost entirely on text prompts, which encode domain-invariant semantics but miss domain-specific visual information needed for precise localization under few-shot supervision. We propose a dual-branch detector that Learns Multi-modal Prototypes, dubbed LMP, by coupling textual guidance with visual exemplars drawn from the target domain. A Visual Prototype Construction module aggregates class-level prototypes from support RoIs and dynamically generates hard-negative prototypes in query images via jittered boxes, capturing distractors and visually similar backgrounds. In the visual-guided branch, we inject these prototypes into the detection pipeline with components mirrored from the text branch as the starting point for training, while a parallel text-guided branch preserves open-vocabulary semantics. The branches are trained jointly and ensembled at inference by combining semantic abstraction with domain-adaptive details. On six cross-domain benchmark datasets and standard 1/5/10-shot settings, our method achieves state-of-the-art or highly competitive mAP.

\end{abstract}

%% file: sec/1_intro.tex
\section{Introduction}
\label{sec:intro}
\begin{figure}
  \centering
  \includegraphics[width=0.99\linewidth]{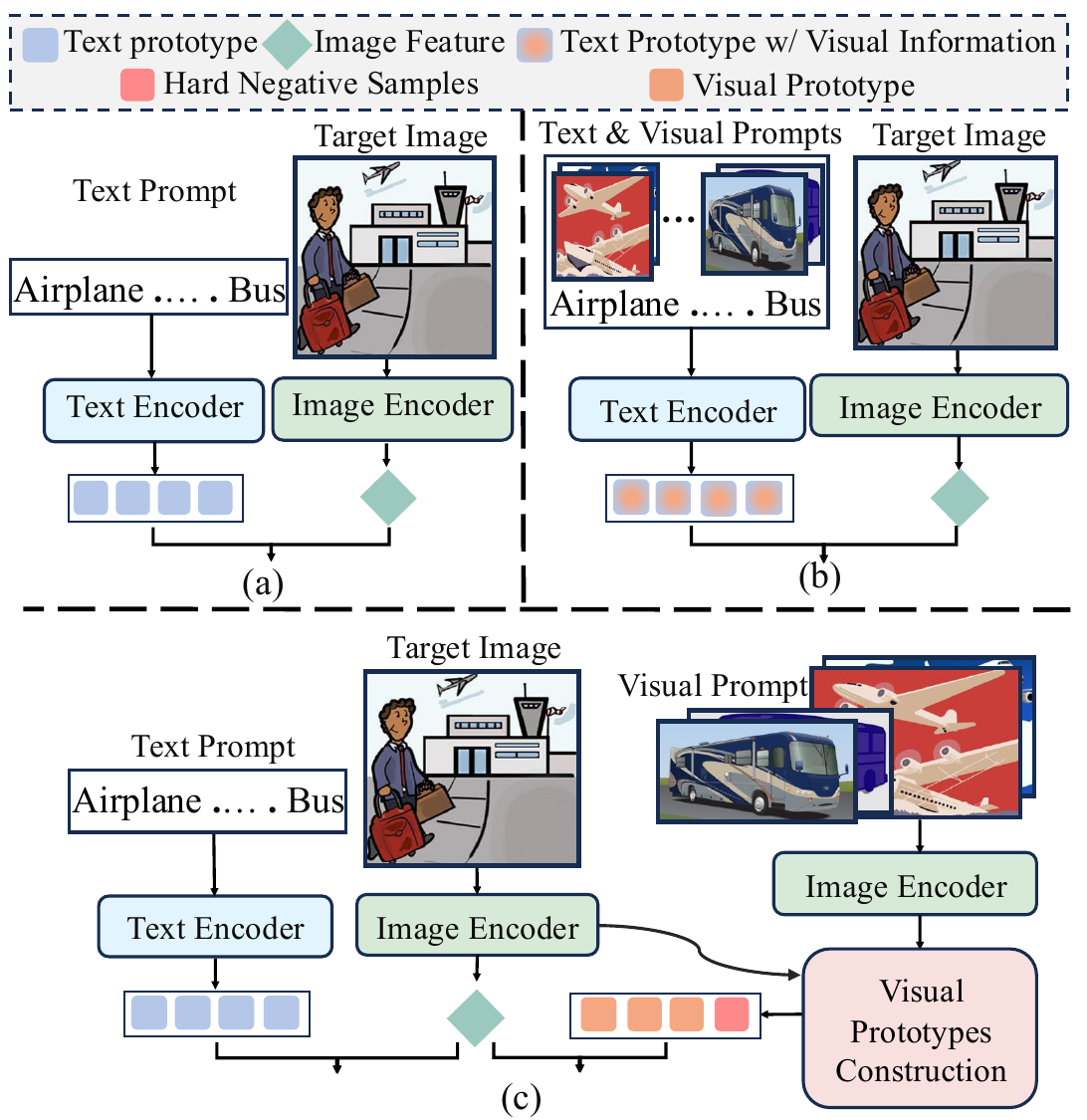}
  \caption{Text vs. visual prompting for cross-domain few-shot detection.
(a) Text-only prompts encode high-level semantics but miss target-domain appearance, leading to weak localization under domain shift.
(b) Adding raw visual prompts (support images) enriches semantics but still lacks structured, class-specific guidance.
(c) Our approach constructs compact visual prototypes from support images and injects them alongside text features into the detector, which provides domain-adaptive capability for robust FSOD.}
  \label{fig:short}
\end{figure}
Few-shot object detection (FSOD)~\cite{antonelli2022few,kang2019few,wang2024fine,han2023few} aims to recognize and localize novel object categories with limited labeled examples. The task typically assumes that the base and novel classes are drawn from the same data distribution. However, this assumption often fails in real-world scenarios where domain shifts exist. This motivates the research on Cross-Domain Few-Shot Object Detection (CD-FSOD)~\cite{fu2025ntire,fu2024cross,pan2025enhance,li2025domain,gao2022acrofod}, which aims to transfer detection capability from a source domain to previously unseen target domains with only a few labeled instances of novel classes.

Recent advances in Vision-Language Models (VLMs) have demonstrated remarkable zero-shot transferability through large-scale pre-training on image-text pairs. VLMs such as CLIP~\cite{radford2021learning} and ALIGN~\cite{jia2021scaling} have shown strong generalization to downstream tasks with minimal supervision~\cite{zhang2022tip,zhou2022learning,wei2024benchmarking,zhao2025continual,lim2024dipex,chen2025cluster,lim2024track,chen2025fastedit}. This paradigm naturally extends to object detection, where open-vocabulary detectors like GLIP~\cite{li2022grounded} and GroundingDINO~\cite{liu2024grounding} leverage text prompts to achieve flexible object-text alignment. Such VLM-based detectors have shown promise for CD-FSOD, as their rich pre-trained representations can potentially adapt to novel domains with limited target-domain annotations~\cite{fu2025ntire,pan2025enhance,li2025domain}.

However, VLM-based detectors often rely on text prompts as the only guidance. Although text encodes high-level semantics that are relatively invariant across domains, it does not capture how categories appear under domain shift. For instance, the text ``airplane" provides the same semantic representation regardless of the visual domain. In contrast, the visual features of the same semantic category vary across different domains. 
Natural images exhibit different viewpoints and complex backgrounds, remote-sensing imagery presents near-nadir perspectives and uniform scales, while cartoon domains are characterized by stylized linework and exaggerated shape deformations. 
These domain-specific visual characteristics are not explicitly captured by text-only representations. 
The limitation becomes even challenging in a few-shot setting. With only a few labelled examples, the detector cannot learn domain-specific appearance through data alone. Text guidance is insufficient to bridge the gap.

Thus, the motivation of our work is that visual prototypes extracted from target-domain support images can provide the missing domain information that text prototypes lack. 
Unlike text prototypes that encode semantically what an object is, visual prototypes encode how objects appear in a specific domain. This allows for capturing fine-grained patterns such as canonical viewpoints, illumination characteristics, and stylistic rendering. Importantly, the few-shot support set, despite containing limited instances, already embodies representative visual patterns of the target domain that can be leveraged to \textbf{construct domain-adaptive prototypes}. Furthermore, achieving robust discrimination requires the model to learn not only positive visual patterns from support images but also to \textbf{explicitly model negative patterns}. Specifically, the model should capture domain-specific background regions and distractors that exhibit visual similarity to target objects, as these constitute the primary sources of false positives in CD-FSOD scenarios.

To this end, we propose a dual-branch framework that Learns Multi-modal Prototypes (LMP) for Cross-Domain Few-Shot Object Detection. Built upon GroundingDINO~\cite{liu2024grounding}, our method preserves the original text-guided branch to maintain strong semantic understanding and open-vocabulary capability, while introducing a parallel visual-guided branch to inject domain-specific visual prototypes. Central to the visual-guided branch is a Visual Prototype Construction module that learns from both class-level prototypes and hard negative examples. The class-level prototypes are aggregated from support image RoI features, which contain representative visual characteristics of target categories within the specific domain. Critically, hard negative prototypes are dynamically generated during training by applying random jittering to the ground-truth bounding boxes on query images. They create perturbed regions that capture domain-specific distractors and visually confusing backgrounds. By incorporating both positive and negative prototypes into a unified representation and jointly training them, our model learns to discriminate between true objects and false positives without requiring additional contrastive objectives. 
Building upon constructed visual prototypes, the visual-guided branch employs a detection pipeline that performs prototype-guided feature enhancement, similarity-based query selection, and iterative refinement through a dedicated visual decoder. Throughout this pipeline, visual prototypes actively guide both feature representation and spatial localization, while the text-guided branch operates in parallel to provide semantic constraints, with both branches collaborating during inference through ensemble predictions.

The main contributions of this paper are summarized as follows:
\begin{itemize}
    \item A dual-branch CD-FSOD framework that integrates textual and visual guidance. The text branch maintains open-vocabulary semantics, and the visual branch injects domain-adaptive appearance via visual prototypes. Predictions are ensembled at inference.

    \item A Visual Prototype Construction module that unifies \emph{class-level} prototypes (from supports) with \emph{hard negative} prototypes (from query jittering), explicitly modelling domain-specific distractors such as visually similar backgrounds and partial overlaps.

    
    \item Extensive experiments on ArTaxOr, Clipart1k, DIOR, DeepFish, NEU-DET, UODD and 1/5/10-shot settings show state-of-the-art or highly competitive performance.

\end{itemize}

%% file: sec/2_formatting.tex
\section{Related work}
\label{sec:formatting}
\textbf{Few-Shot Objection Detection (FSOD).}
Existing FSOD approaches ~\cite{kohler2023few,wu2020multi,li2021transformation,demirel2023meta,zhang2022time,bulat2023fs,xin2024few,zhang2024weakly} broadly fall into two groups, including Meta-learning-based and transfer-learning-based methods.  Meta-learning-based methods optimize the model at the task level via episodic training across multiple few-shot tasks, thereby enabling rapid adaptation and improved generalization to unseen classes (e.g., Meta R-CNN~\cite{yan2019meta}, Meta-DETR~\cite{zhang2022meta}), FCT~\cite{han2022few}. Transfer-learning methods initialize detectors with weights pretrained on base classes and then fine-tune them on the few-shot novel classes (e.g., TFA~\cite{wang2020frustratingly}, FSCE~\cite{sun2021fsce}, DeFRCN~\cite{qiao2021defrcn}). Recently, Han et al.~\cite{han2024few} leveraged modern foundation models (DINOv2~\cite{oquab2023dinov2}) for visual feature extraction and exploited the in-context learning capabilities of large language models (LLMs) for contextualized few-shot proposal classification in FSOD. Despite substantial progress, most FSOD methods still assume that the training/source and the testing/target data are drawn from the same domain.

\noindent\textbf{Cross-Domain Few-Shot Objection Detection (CD-FSOD).}
Existing CD-FSOD approaches ~\cite{fu2025ntire,fu2024cross,pan2025enhance,li2025domain,Huang_2025_CVPR,meng2025cdformer,lee2022rethinking,gao2022acrofod,xiong2023cd,liudon,zhao2022oa} can be broadly organized into closed-source and open-source settings~\cite{fu2025ntire}. Closed-source methods restrict the accessible source-domain training data to a fixed dataset. For example, Fu et al.~\cite{fu2024cross} introduce the CD-FSOD benchmark and propose CD-ViTO under a setting where the source data are strictly limited to MS-COCO~\cite{lin2014microsoft} only. In contrast, open-source CD-FSOD methods are designed to leverage the capabilities of foundation models. 
ETS~\cite{pan2025enhance} combines mixed image augmentation with a grid-based sub-domain search strategy via pretrained GroundingDINO~\cite{liu2024grounding} to efficiently search for optimal sub-domains within a broad domain space. 
Domain-RAG~\cite{li2025domain} develops a retrieval-guided compositional image generation framework built upon the principle of ``fix the foreground and adapt the background" for boosting CD-FSOD. Empirically, such open-source approaches often outperform their closed-source counterparts under matched target domains.

\noindent\textbf{Vision Language Models for CD-FSOD.}
VLMs leverage web-scale image-text pairs with contrastive objectives to align separately encoded image and text representations in a joint embedding space. CLIP~\cite{radford2021learning} and ALIGN~\cite{jia2021scaling} achieve effective image-level cross-modal alignment and enable strong transfer to image classification and text-image retrieval tasks. Inspired by these works, object-level visual representations are highly desired. GLIP~\cite{li2022grounded} and GroundingDINO~\cite{liu2024grounding} reformulate object detection as phrase grounding through deep cross-modal fusion. GroundingDINO further extends this framework by incorporating the DINO detection backbone and enhanced feature fusion mechanisms for improved region-text alignment. However, these models use only text for querying objects, which struggles with fine-grained distinctions as similar categories occupy nearby regions in text embedding space.
MQ-Det~\cite{xu2023multimodal} integrates class-wise cross-attention layers within the text encoder to re-weight text queries with visual exemplars.  
VisTex-OVLM~\cite{wu2025visual} projects visual exemplars into textualized visual tokens to guide Object-level Vision-Language Models (OVLMs) alongside text prompts. 
Unlike prior works that integrate visual information into text encoders through feature fusion or projection, we propose a dual-branch architecture where a separate visual-guided branch learns structured class-level prototypes from support images and dynamically generates hard negative prototypes via ground-truth box jittering. Our method explicitly addresses both domain adaptation and confusable context in CD-FSOD.





%% file: sec/3_finalcopy.tex
\vspace{-10pt}
\section{Learning Multi-Modal Prototypes}
\begin{figure*}
  \centering
  \includegraphics[width=0.99\linewidth]{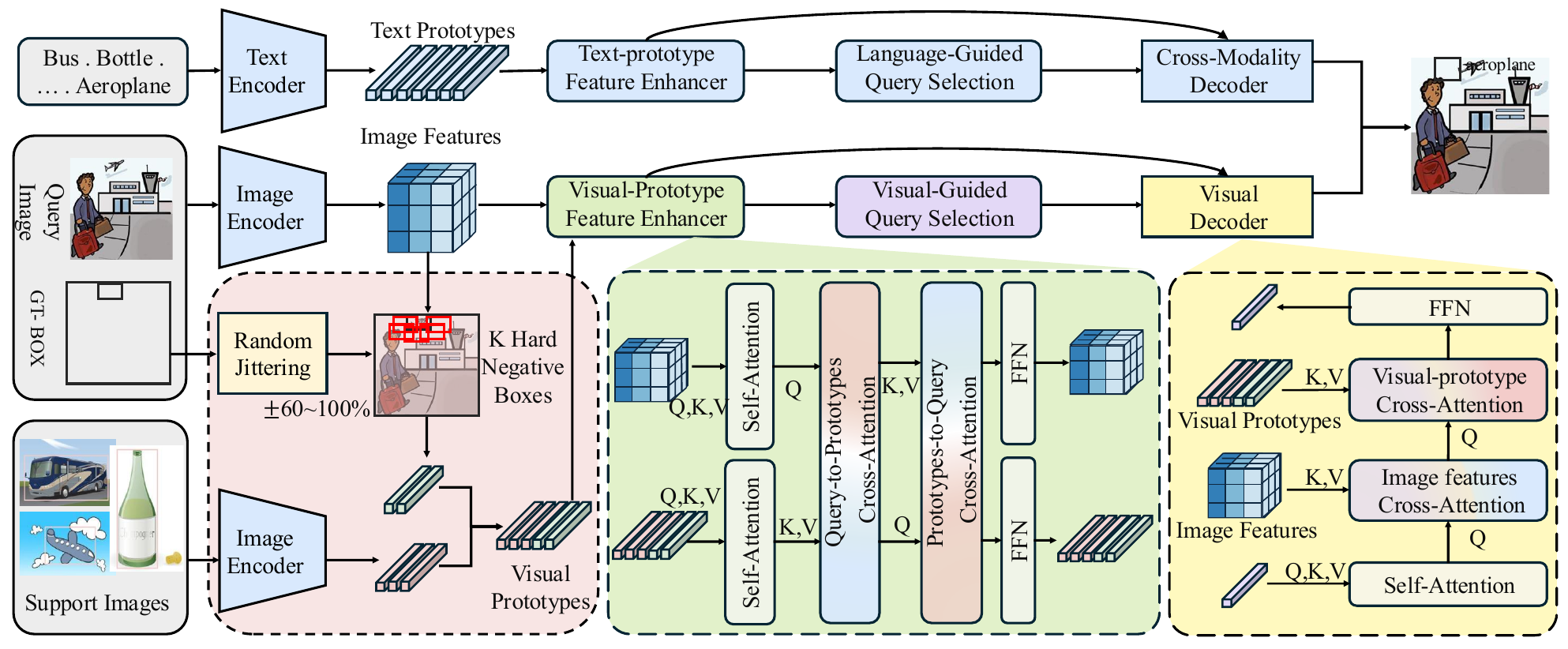}
  \caption{Overview of the proposed LMP framework for CD-FSOD. From a few labeled support images, we build class-level visual prototypes and, for each ground-truth in a query image, sample K hard-negative boxes via random jittering. A visual-guided branch injects these prototypes into the detection pipeline, while a text-guided branch preserves open-vocabulary semantics. The two branches are trained jointly and ensembled at inference, coupling domain-invariant text features with target-domain appearance for robust few-shot detection.}
  \label{fig:short}
\end{figure*}
\subsection{Task definition}
Cross-Domain Few-Shot Object Detection (CD-FSOD) considers two distinct data domains with different distributions from a source domain and a target domain. In the target domain, the labeled set follows a C-way K-shot protocol, where the support set consists of K annotated instances from each of the C novel classes for few-shot learning, while the query set Q is used for prediction. The goal of CD-FSOD is to transfer detection capability from the source domain to previously unseen target domains and  detect objects in query images.
We adopt the open-source setting introduced in the 1st CD-FSOD Challenge~\cite{fu2025ntire}, which aims to leverage the foundation models pre-trained on large-scale datasets and explore the potential of foundation models. Instead of accessing source-domain training data, we directly fine-tune the foundation model GroundingDINO~\cite{liu2024grounding} with the few-shot data of the target domains.




\subsection{Motivation}
In CD-FSOD, a detector must generalize across appearance shifts, such as style, texture, lighting, capture conditions, while learning from only a handful of labelled examples per novel class. Open-vocabulary detectors alleviate the label bottleneck by using text prompts, but in CD-FSOD we observe two systematic failure modes: (i) \emph{semantic–appearance mismatch}: text prototypes capture class meaning yet ignore target-domain cues such as rendering style or background texture, leading to poor localisation, and (ii) \emph{confusable context}: with very few positives, visually similar background or near-object regions dominate training and trigger false positives. Simply adding raw support images as visual prompts helps little because unstructured features mix class evidence with incidental context and do not explicitly model hard negatives.

Motivated by these observations, we propose to learn multi-modal prototypes for CD-FSOD with the following considerations:
\begin{itemize}\setlength{\itemsep}{2pt}
    \item \textbf{Domain conditioning without losing openness.} We preserve a text-guided branch so that prompts remain free at inference, but augment it with target-domain visual prototypes that adapt the detector to new appearances.
    \item \textbf{Structured, class-specific visual guidance.} We summarize K-shot supports into compact \emph{visual prototypes} that represent class evidence rather than entire images, and inject them where the detector makes decisions (query selection and decoding).
    \item \textbf{Explicit modelling of confusions.} Construct {hard-negative prototypes} around each ground-truth, e.g., jittered boxes, so the classifier learns to separate objects from their most frequent distractors without extra contrastive losses.
\end{itemize}



\subsection{Visual Prototype Construction}
\noindent\textbf{Class-level Prototypes.} 
Under the $C$-way $K$-shot setting, each support image $I_s$ may contain multiple instances. For every annotated instance $i$ of class $c\!\in\!\{1,\ldots,C\}$ we extract a pooled RoI
feature from the lowest pyramid level of the image encoder via RoIAlign, ${f}_{c,i}^{(0)}\!\in\!\mathbb{R}^{D_I\times H\times W}$, and apply global average pooling (GAP). The class prototype is the mean of its instance descriptors:
\[
\mathbf{p}_c=\frac{1}{|\mathcal{I}_c|}\sum_{i\in\mathcal{I}_c}\operatorname{GAP}\!\big({f}_{c,i}^{(0)}\big)
\in\mathbb{R}^{D_I},
\]
where $\mathcal{I}_c$ indexes all support instances of class $c$.
Stacking all classes yields
\[
\mathbf{P}_{\mathrm{cls}}
=\big[\mathbf{p}_1;\mathbf{p}_2;\ldots;\mathbf{p}_C\big]\in\mathbb{R}^{C\times D_I}.
\]
All prototypes are $\ell_2$-normalized.


\noindent \textbf{Hard Negative Prototypes.} 
To model confusing background near each query ground truth, we generate negatives by
jittering their GT boxes. For the $j$-th GT box $b_j{=}(x_j,y_j,w_j,h_j)$, we sample
$N$ perturbed boxes:
\begin{equation}
\begin{aligned}
\tilde{b}^{(n)}_j=\mathcal{J}(b_j; s^{(n)},\Delta^{(n)}),\quad
s^{(n)}\!\sim\!\mathcal{U}[0.6,1.0],\; \\
\Delta^{(n)}\!\sim\!\mathcal{U}\!\big([{-}0.2w_j,0.2w_j]\times[{-}0.2h_j,0.2h_j]\big),
\end{aligned}
\end{equation}
and keep those with $\operatorname{IoU}(\tilde{b}^{(n)}_j,b_j)\!\in\![0.1,0.5]$.
Each negative prototype is obtained from the lowest-level query feature map
$f_q^{(0)}$ by RoIAlign+GAP:
\begin{equation}
\mathbf{p}^{(n)}_{\mathrm{neg},j}
=\operatorname{GAP}\!\left(\operatorname{RoIAlign}\!\left(f_q^{(0)},\tilde{b}^{(n)}_j\right)\right)
\in\mathbb{R}^{D_I}.
\end{equation}
The set for box $j$ is
$\mathbf{P}^{(j)}_{\mathrm{neg}}=\{\mathbf{p}^{(n)}_{\mathrm{neg},j}\}_{n=1}^{N}$,
and the union over all $J$ GT boxes is
$\mathbf{P}_{\mathrm{neg}}=\bigcup_{j=1}^{J}\mathbf{P}^{(j)}_{\mathrm{neg}}$.
We normalize negatives as well.


\noindent\textbf{Visual Prototypes.} 
For the current query image, we concatenate class and hard-negative prototypes to form the
visual token matrix:
\[
\mathbf{V}=\big[\;\mathbf{P}_{\mathrm{cls}}\;;\;\mathbf{P}_{\mathrm{neg}}\;\big]
\in\mathbb{R}^{N_V\times D_I}, N_V=C+\sum_{j=1}^{J}\!|\mathbf{P}^{(j)}_{\mathrm{neg}}|.
\]
This sequence provides compact, class-specific evidence from support images together with
query-conditioned distractors from jittered boxes, and is consumed by the visual
feature enhancer and the visual decoder in the subsequent subsections.


\subsection{Visual Prototype Refinement}
We reuse the same transformer feature-enhancer and decoder architectures as the language branch, but drive them with the visual prototype sequence\(\mathbf{V}\in\mathbb{R}^{N_V\times D_I}\). To stabilize training, weights are copied from the language branch at the start, and all parameters are then jointly optimized.

\noindent \textbf{Visual-Prototype Feature Enhancer.}\\
Let \(\{f_q^{(\ell)}\in\mathbb{R}^{D_I\times H_\ell\times W_\ell}\}_{\ell=0}^{L}\) be multi-scale image features and
\(\mathbf{X}_I=\mathrm{Concat}_\ell\,\mathrm{Flatten}(f_q^{(\ell)})\in\mathbb{R}^{N_I\times D_I}\) the tokenized image features.
The feature enhancer consists of six self-attention and cross-attention layers:
\[
\begin{aligned}
\tilde{\mathbf{X}}_I = \mathbf{X}_I + \mathrm{SelfAttn}(\mathbf{X}_I), \quad \tilde{\mathbf{V}}   = \mathbf{V}   + \mathrm{SelfAttn}(\mathbf{V}),\\
\mathbf{X}'_I   = \tilde{\mathbf{X}}_I + \mathrm{CrossAttn}\!\left(\tilde{\mathbf{X}}_I,\, \tilde{\mathbf{V}}\right), \\
\mathbf{V}'     = \tilde{\mathbf{V}}   + \mathrm{CrossAttn}\!\left(\tilde{\mathbf{V}},\, \tilde{\mathbf{X}}_I\right),
\end{aligned}
\]
followed by FFN. 
The outputs are adapted prototypes \(\mathbf{V}'\) and  image tokens \(\mathbf{X}'_I\).


\noindent \textbf{Visual-Guided Query Selection.}
Given the prototype-aware image tokens $\mathbf{X}'_I$  and the adapted visual prototype sequence $\mathbf{V}'$, we score each image token by its maximum cosine similarity to any guidance token, then take the Top–$N_q$ indices to initialize queries. With normalized tokens,
let $\mathbf{S}=\mathbf{X}'_I{\mathbf{V}'}^{\!\top}\in\mathbb{R}^{N_I\times N_V}, s_i=\max_{v\le N_V}\mathbf{S}_{i,v}$.
We select:
\[
\mathcal{I}^{\mathrm{vis}}_{N_q}=\operatorname*{TopK}_{N_q}\big(\{s_i\}_{i=1}^{N_I}\big),\quad N_q=900.
\]
For each selected index $i\!\in\!\mathcal{I}^{\mathrm{vis}}_{N_q}$, we form a mixed query seed consisting of a positional component from a dynamic anchor $a_i=(x_i,y_i,w_i,h_i)$ predicted at
the corresponding pyramid level and embedded by $\mathbf{e}_{\mathrm{pos}}(a_i)$, and a learnable
content component $\mathbf{e}_{\mathrm{cnt}}$. The initial query is:
\[
\mathbf{q}^{(0)}_i=\mathbf{e}_{\mathrm{pos}}(a_i)\ \oplus\ \mathbf{e}_{\mathrm{cnt}}\in\mathbb{R}^{d}.
\]
These seeds provide both reference anchors and content embeddings for the decoder.


\noindent \textbf{Visual Decoder.}
The decoder mirrors the cross-modality decoder where each layer applies (1) self-attention on queries, (2) multi-scale \emph{deformable} cross-attention to image features, (3) cross-attention to the adapted visual prototypes $\mathbf{V}'$, followed by (4) an FFN, with norm and residual connections throughout. 
Prediction heads at each layer output class logits and box
deltas, and reference boxes are updated iteratively via $b^{t+1}=b^{t}+\Delta b^{t}$. Classification uses
prototype-aligned scoring by cosine similarity between projected query embeddings and class visual prototypes.

\subsection{Text-Guided Prototype Refinement}
We retain GroundingDINO’s text branch to preserve open-vocabulary detection. Let $\mathcal{T}=\{t_c\}_{c=1}^{C}$ be the class names. The text encoder yields $d$-dimensional embeddings, which we project to the detector width $D_I$:
\[
\mathbf{P}_{\text{text}}=\big[\mathbf{p}^{\text{text}}_c\big]_{c=1}^{C},
\tilde{\mathbf{P}}_{\text{text}}=\mathrm{Norm}\!\left(\frac{\mathbf{P}_{\text{text}} W_t}{\tau_t}\right)\in\mathbb{R}^{C\times D_I}.
\]
We then follow the same three stages as in the visual pathway but swap visual prototypes for $\tilde{\mathbf{P}}_{\text{text}}$:
(i) text-prototype feature enhancement, where image tokens attend to $\tilde{\mathbf{P}}_{\text{text}}$;
(ii) language-guided query selection, which scores image tokens by their maximum cosine similarity to $\tilde{\mathbf{P}}_{\text{text}}$ and selects the Top-$N_q$  to seed queries; and
(iii) cross-modality decoding, which 
produces text-guided detections.
At inference, this branch is ensembled with the visual branch to couple text branch with domain-adaptive appearance.


\subsection{Optimization}
We adopt a dual-branch supervision that trains the text- and visual-guided branches jointly with one-to-one Hungarian matching. Let each decoder layer output $N_q$ predictions of class logits and boxes, we supervise the last layer and add auxiliary losses to intermediate layers.

\noindent \textbf{Logits and targets.}
For the text branch, a projection maps query embeddings to the prototype space and class logits are computed by cosine similarity to text prototypes $\tilde{\mathbf{P}}_{\text{text}}\!\in\!\mathbb{R}^{C\times D_I}$:
$
z^{\text{text}}_{i,c}=\frac{\alpha_{\text{txt}}}{\tau_{\text{txt}}}\,
\cos\!\big(\mathbf{W}_{\text{cls}}\mathbf{q}^{\text{text}}_i,\ \tilde{\mathbf{p}}^{\text{text}}_c\big).
$
For the visual branch, logits are computed against the class \emph{visual} prototypes $\mathbf{P}_{\mathrm{cls}}$ and hard negatives $\mathbf{P}_{\mathrm{neg}}$ do not form categories but expand the background target through the attention path:
$
z^{\text{vis}}_{i,c}=\frac{\alpha_{\text{vis}}}{\tau_{\text{vis}}}\,
\cos\!\big(\mathbf{W}_{\text{cls}}\mathbf{q}^{\text{vis}}_i,\ \mathbf{p}_c\big),\qquad c=1,\ldots,C.
$
All tokens are $\ell_2$-normalized before cosine, and $\tau_{\{\cdot\}}$ are fixed temperatures.
For a matched query $i\!\leftrightarrow\!(b^\ast,y^\ast)$ with class $y^\ast\!\in\!\{1,\ldots,C\}$, the focal target is one-hot on $y^\ast$ and zero elsewhere. Hard negatives contribute to the {background} channel by injecting into cross-attention and increase the mass of “non-class” during training, so high scores on distractors are down-weighted by the focal term without an extra contrastive loss.

\noindent \textbf{Branch losses.}
Each branch uses focal classification plus $L_1$ and GIoU for boxes, which are summed over matched queries, with auxiliary losses at intermediate decoder layers:
\[
\mathcal{L}_{\text{text}}
=\lambda_{\text{cls}}\mathcal{L}_{\text{focal}}^{\text{text}}
+\lambda_{\ell_1}\mathcal{L}_{\ell_1}^{\text{text}}
+\lambda_{\text{giou}}\mathcal{L}_{\text{giou}}^{\text{text}},
\]
\[
\mathcal{L}_{\text{visual}}
=\lambda_{\text{cls}}\mathcal{L}_{\text{focal}}^{\text{visual}}
+\lambda_{\ell_1}\mathcal{L}_{\ell_1}^{\text{visual}}
+\lambda_{\text{giou}}\mathcal{L}_{\text{giou}}^{\text{visual}}.
\]

\noindent \textbf{Total loss.}
We combine branches with a scalar $\alpha$:
\[
\mathcal{L}_{\text{total}}=\mathcal{L}_{\text{text}}+\alpha\,\mathcal{L}_{\text{visual}},
\]
where $\alpha$ is coefficient to balance the visual/text prototype supervision.

%% file: sec/4_Experiments.tex
\begin{table*}[t]
\centering
\setlength{\tabcolsep}{4pt}
\caption{Main Results (mAP) on CD-FSOD benchmark under the 1/5/10-shot settings. }
\label{tab:fsod_cross_domain_grouped}
\scalebox{0.89}{
\begin{tabular}
{c l l r c c c c c c c}
\toprule
\textbf{Shots} & \textbf{Method} & \textbf{Venue} & \textbf{Backbone} &
\textbf{ArTaxOr} & \textbf{Clipart1k} & \textbf{DIOR} & \textbf{DeepFish} &
\textbf{NEU-DET} & \textbf{UODD} & \textbf{Average}\\
\midrule

Z/Shot
& GroundingDINO~\cite{liu2024grounding} & ECCV'24 & Swin-B 
& 12.4 & 54.0 & 4.8 & 35.4 & 3.4 & 13.9 &20.7 \\
\midrule
\shotcell{13}{1-shot}

& TFA w/cos~\cite{wang2020frustratingly} &ICML'20& ResNet50 
& 3.1 & - & 8.0 & - & - & 4.4 & / \\

& FSCE~\cite{sun2021fsce} &CVPR'21& ResNet50 
& 3.7 & - & 8.6 &- & - & 3.9 & / \\

& DeFRCN~\cite{qiao2021defrcn}&ICCV'21 & ResNet50 
& 3.6 & - & 9.3 & - & - & 4.5 & / \\


\cmidrule(r){2-11}
& ViTDet-FT~\cite{li2022exploring} &ECCV'22 & ViT-B/14 
& 5.9 & 6.1 & 12.9 & 0.9 & 2.4 & 4.0 & 5.4 \\


& Detic-FT~\cite{zhou2022detecting} &ECCV'22 & ViT-L/14 
& 3.2 & 15.1 & 4.1 & 9.0 & 3.8 & 4.2 & 6.6 \\

& DE-ViT~\cite{zhang2023detect} &CoRL'24 & ViT-L/14 
& 0.4 & 0.5 & 2.7 & 0.4 & 0.4 & 1.5 & 1.0 \\

& CD-ViTO~\cite{fu2024cross} &ECCV'24 & ViT-L/14 
& 21.0 & 17.7 & 17.8 & 20.3 & 3.6 & 3.1 & 13.9 \\
\cmidrule(r){2-11}

& GroundingDINO~\cite{liu2024grounding} &ECCV'24& Swin-B 
& 26.3 & 55.3 & 14.8 & 36.4 & 9.3 & 15.9 & 26.3 \\


& ETS~\cite{pan2025enhance} &CVPRW'25 & Swin-B 
& 28.1 & 55.8 & 12.7 & \textbf{39.3} & 11.7 & 18.9 & 27.8 \\


& Domain-RAG~\cite{li2025domain} &NeurIPs'25 & Swin-B 
& 57.2 & 56.1 & \textbf{18.0} & 38.0 & 12.1& 20.2 & 33.6 \\

& \textbf{LMP (ours)}& CVPR'26 & Swin-B 
& \textbf{58.5} & \textbf{58.6} & 17.2 & 36.6 & \textbf{13.3} & \textbf{21.6} & \textbf{34.3} \\

\midrule

\shotcell{13}{5-shot}

&TFA w/cos~\cite{wang2020frustratingly} &ICML'20& ResNet50 
& 8.8 & - & 18.1 & - & - & 8.7 & / \\

& FSCE~\cite{sun2021fsce} &CVPR'21 & ResNet50 
& 10.2 & - & 18.7 & - & - & 9.6 & / \\

& DeFRCN~\cite{qiao2021defrcn}&ICCV'21 & ResNet50 
& 9.9 & - & 18.9 & - & - & 9.9 & / \\


\cmidrule(r){2-11}

& ViTDet-FT~\cite{li2022exploring} &ECCV'22 & ViT-B/14 
& 20.9 & 23.3 & 23.3 & 9.0 & 13.5 & 11.1 & 16.9 \\


& Detic-FT~\cite{zhou2022detecting} &ECCV'22 & ViT-L/14 
& 8.7 & 20.2 &12.1 & 14.3 &14.1 &10.4 & 13.3 \\
 
&DE-ViT~\cite{zhang2023detect} &CoRL'24& ViT-L/14 
& 10.1 & 5.5 & 7.8 & 2.5 & 1.5 & 3.1 & 5.1 \\

& CD-ViTO~\cite{fu2024cross} &ECCV'24 & ViT-L/14 
& 47.9 & 41.1 & 26.9 & 22.3 & 11.4 & 6.8 & 26.1 \\
\cmidrule(r){2-11}
& GroundingDINO~\cite{liu2024grounding}&ECCV'24 & Swin-B 
& 68.4 & 57.6 & 29.6 & 41.6 & 19.7 & 25.6 & 40.4 \\


& ETS~\cite{pan2025enhance} &CVPRW'25 & Swin-B 
& 64.5 & 59.7 & 29.3 & 42.1 & 23.5 & 27.7 & 41.1 \\


& Domain-RAG~\cite{li2025domain} &NeurIPs'25  & Swin-B 
& 70.0 & 59.8 & \textbf{31.5} & 43.8 & 24.2 & 26.8 & 42.7 \\

& \textbf{LMP (ours)} & CVPR'26 & Swin-B 
& \textbf{75.0} & \textbf{60.1} & 31.3 & \textbf{43.9} & \textbf{25.1} & \textbf{28.3} & \textbf{44.0} \\
\midrule

\shotcell{13}{10-shot}

&TFA w/cos~\cite{wang2020frustratingly} &ICML'20 & ResNet50 
& 14.8 & - & 20.5 & - & - & 11.8 & / \\

& FSCE~\cite{sun2021fsce} &CVPR'21 & ResNet50 
& 15.9 & - & 21.9 & - & - & 12.0 & / \\

& DeFRCN~\cite{qiao2021defrcn}&ICCV'21 & ResNet50 
& 15.5 & - & 22.9 & - & - & 12.1 & / \\

\cmidrule(r){2-11}
& ViTDet-FT~\cite{li2022exploring} &ECCV'22 & ViT-B/14 
& 23.4 & 25.6 & 29.4 & 6.5 & 15.8 & 15.6 & 19.4 \\


& Detic-FT~\cite{zhou2022detecting} &ECCV'22& ViT-L/14 
& 12.0 & 22.3 & 15.4 & 17.9 & 16.8 & 14.4 & 16.5 \\

& DE-ViT~\cite{zhang2023detect} &CoRL'24 & ViT-L/14 
& 9.2 & 11.0 & 8.4 & 2.1 & 1.8 & 3.1 & 5.9 \\

& CD-ViTO~\cite{fu2024cross} &ECCV'24 & ViT-L/14 
& 60.5 & 44.3 & 30.8 & 22.3 & 12.8 & 7.0 & 29.6 \\
\cmidrule(r){2-11}
& GroundingDINO~\cite{liu2024grounding}&ECCV'24 & Swin-B 
& 73.0 & 58.6 & 37.2 & 38.5 & 25.5 & 30.3 & 43.9 \\

& ETS~\cite{pan2025enhance} &CVPRW'25 & Swin-B 
& 70.6 & 60.8 & 37.5 & 42.8 & 26.1 & 28.3 & 44.4 \\


& Domain-RAG~\cite{li2025domain} &NeurIPs'25  & Swin-B 
& 73.4 & 61.1 & 39.0 & 41.3 & \textbf{26.3} & 31.2 & 45.4 \\

& \textbf{LMP (ours)} & CVPR'26 & Swin-B 
& \textbf{75.1} & \textbf{62.6} & \textbf{40.4} & \textbf{44.4} & 25.7 & \textbf{31.6} &  \textbf{46.6}\\
\bottomrule
\end{tabular}}
\vspace{-5pt}
\end{table*}

\section{Experiments}
\noindent\textbf{Datasets.}
We follow the CD-ViTO benchmark~\cite{fu2024cross} to evaluate the performance of our LMP across six target domains that span diverse visual characteristics, including ArTaxOr~\cite{drange2019arthropod} for photorealistic images, Clipart1k~\cite{inoue2018cross} for cartoon illustrations, DIOR ~\cite{li2020object} for aerial photography, DeepFish ~\cite{saleh2020realistic} and UODD~\cite{jiang2021underwater} for underwater imagery, and NEU-DET~\cite{song2013noise} for industrial defect inspection. Performance is measured using mean Average Precision (mAP) under 1-shot, 5-shot, and 10-shot settings, consistent with the benchmark's standard protocol.

\noindent\textbf{Implementation.} 
We use pretrained GroundingDINO~\cite{liu2024grounding} with Swin Transformer-Base (Swin-B)~\cite{liu2021swin} and BERT-base~\cite{devlin2019bert} as our baseline detector. The model is fine-tuned in two stages. In the first stage, only the visual-guided branch is optimized using the visual loss $\mathcal{L}_{\text{visual}}$ with 900 queries. In the second stage, the visual-guided branch and the text-guided branch are trained jointly with the full objective $\mathcal{L}_{\text{total}}$, where each branch selects 900 queries and the loss weighting factor $\alpha$ is set to 1.0 to balance visual and text supervision. 
 We use AdamW~\cite{loshchilov2017decoupled} with a learning rate of 1e-4 and a weight decay of 1e-4, while the backbone learning rate is set to 1e-5. All experiments are conducted on a single NVIDIA GeForce RTX 3090 GPU. 


\subsection{Main Comparison Results}
The 1/5/10-shot results on the six target domains are summarized in Table \ref{tab:fsod_cross_domain_grouped}. We compare our method with representative few-shot detectors, including TFA w/cos~\cite{wang2020frustratingly}, FSCE~\cite{sun2021fsce}, DeFRCN~\cite{qiao2021defrcn}, ViTDet-FT~\cite{zhang2023detect}, Detic-FT~\cite{zhou2022detecting}, DE-ViT~\cite{zhang2023detect},and CD-ViTO~\cite{fu2024cross}, using the numbers reported in CD-ViTO~\cite{fu2024cross}. 
In addition, we compare against GroundingDINO-based methods, ETS~\cite{pan2025enhance}, Domain-RAG~\cite{li2025domain} and a vanilla fine-tuned GroundingDINO~\cite{liu2024grounding} baseline, using the numbers reported in Domain-RAG~\cite{li2025domain}. 

\begin{figure*}
  \centering
  \begin{subfigure}[t]{0.40\textwidth}
    \includegraphics[width=\linewidth]{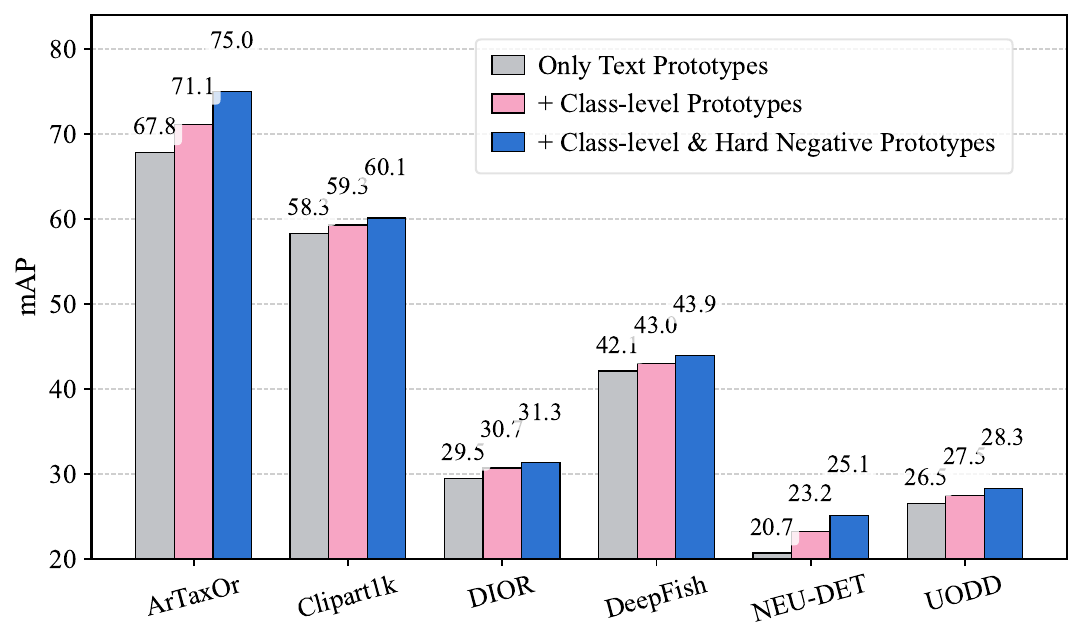}
    \caption{Ablation Study}
    \label{fig:ablation}
  \end{subfigure}\hfill
  \begin{subfigure}[t]{0.58\textwidth}
    \includegraphics[width=\linewidth]{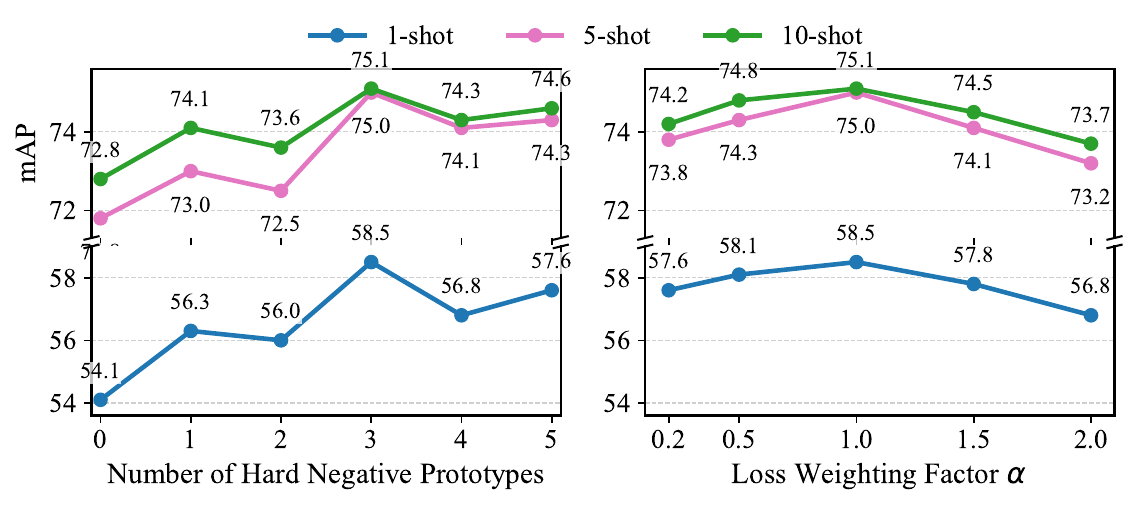}
    \caption{Hyper-parameter sensitivity analysis on the ArTaxOr dataset.}
    \label{fig:hyper-parameter}
  \end{subfigure}\hfill
  \caption{(a) Ablation study results reported on six target-domain datasets, 5-shot. (b) Impact of (left) the number of hard negative prototypes and (right) the loss weighting factor $\alpha$ on detection performance across 1/5/10-shot settings on ArTaxOr dataset.  }
  \label{fig:short}
\end{figure*}

Overall, our LMP  achieves state-of-the-art or highly competitive performance across all settings. Averaged over the six target domains, LMP improves upon the GroundingDINO baseline by 8.0, 3.6, and 2.1 mAP in the 1-, 5-, and 10-shot settings, respectively.
Beyond the averaged performance gains, we highlight two key observations. First, we obtain significant gains on coarse-label datasets. ArTaxOr uses coarse taxonomic labels, e.g., ``Coleoptera", ``Lepidoptera", that convey minimal visual information about shape, texture, or color.  Text-only prompts offer weak guidance for distinguishing visually similar species. By constructing visual prototypes, LMP injects domain-specific appearance supervision that texts lack, which results in substantial improvements. Second, LMP shows effectiveness under extreme data scarcity. The largest average gains occur in the 1-shot regime with 8.0 mAP improvement, with margins decreasing as shot increases.  This demonstrates that multi-modal prototypes prove effective when only a single annotated instance per class is available.

\begin{figure}
  \centering
  \includegraphics[width=0.95\linewidth]{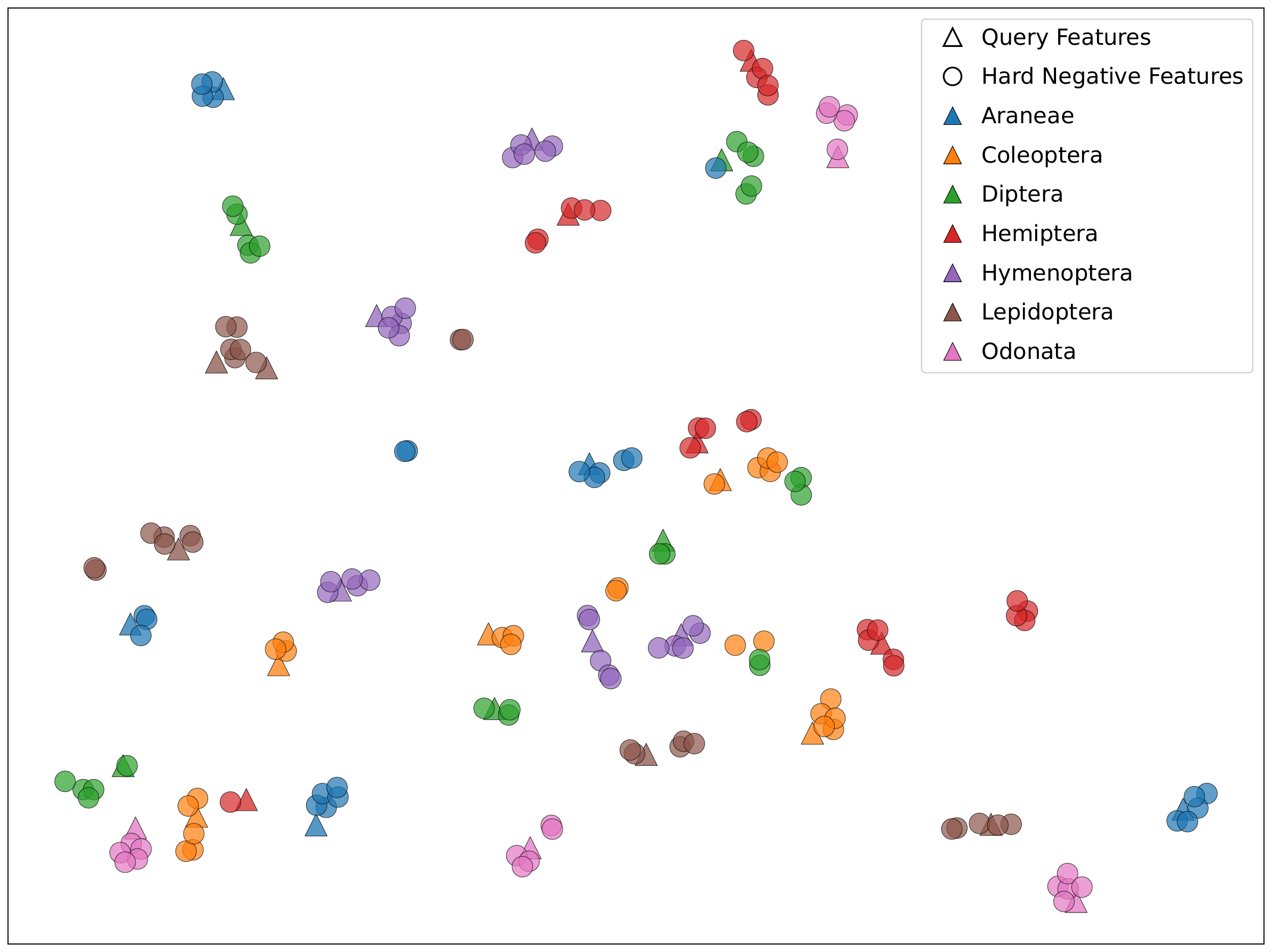}
\caption{t-SNE visualization of prototype embeddings in ArTaxOr. Triangles are query features of detections and circles are hard-negative features mined around ground-truth boxes. Colors denote different classes. Hard negatives cluster along decision boundaries, while query features form different class groups, which shows how the visual branch separates confusing objects.}
  \label{fig:tsne}
\end{figure}

\subsection{More Analysis}
\noindent\textbf{Ablation Study.}
We conduct ablation experiments to examine the impact of different prototype types under the 5-shot setting across six target domains, as shown in Figure \ref{fig:ablation}. We compare three configurations. First, using text prototypes only, we fine-tune GroundingDINO as the baseline. Text embeddings in GroundingDINO lack target-domain visual understanding. Second, we add class-level visual prototypes to the visual-guided branch and achieve consistent improvements over the text-only baseline across all datasets. Finally, we further incorporate hard negative prototypes alongside class-level prototypes and achieve optimal performance across all datasets. This demonstrates that hard negative prototypes are essential for handling data scarcity and visual ambiguity in cross-domain detection.

\noindent\textbf{Hyperparameter Sensitivity Analysis.} 
We investigate the impact of two critical hyperparameters, which are the number of hard negative prototypes $N$ and the loss weighting factor $\alpha$. As shown in Figure \ref{fig:hyper-parameter}, employing three hard negative prototypes per ground-truth achieves optimal detection accuracy, with particularly substantial gains in the 1-shot regime. However, increasing $N$ to 5 causes slight performance degradation. 
The right panel of Figure \ref{fig:hyper-parameter} examines the contribution balance between visual and text prototypes supervision via $\alpha$. The best performance occurs at $\alpha=1.0$ across all shot settings. This indicates that equal weighting optimally integrates domain-adaptive visual cues with semantic abstraction.

\begin{figure*}
  \centering
  \includegraphics[width=0.99\linewidth]{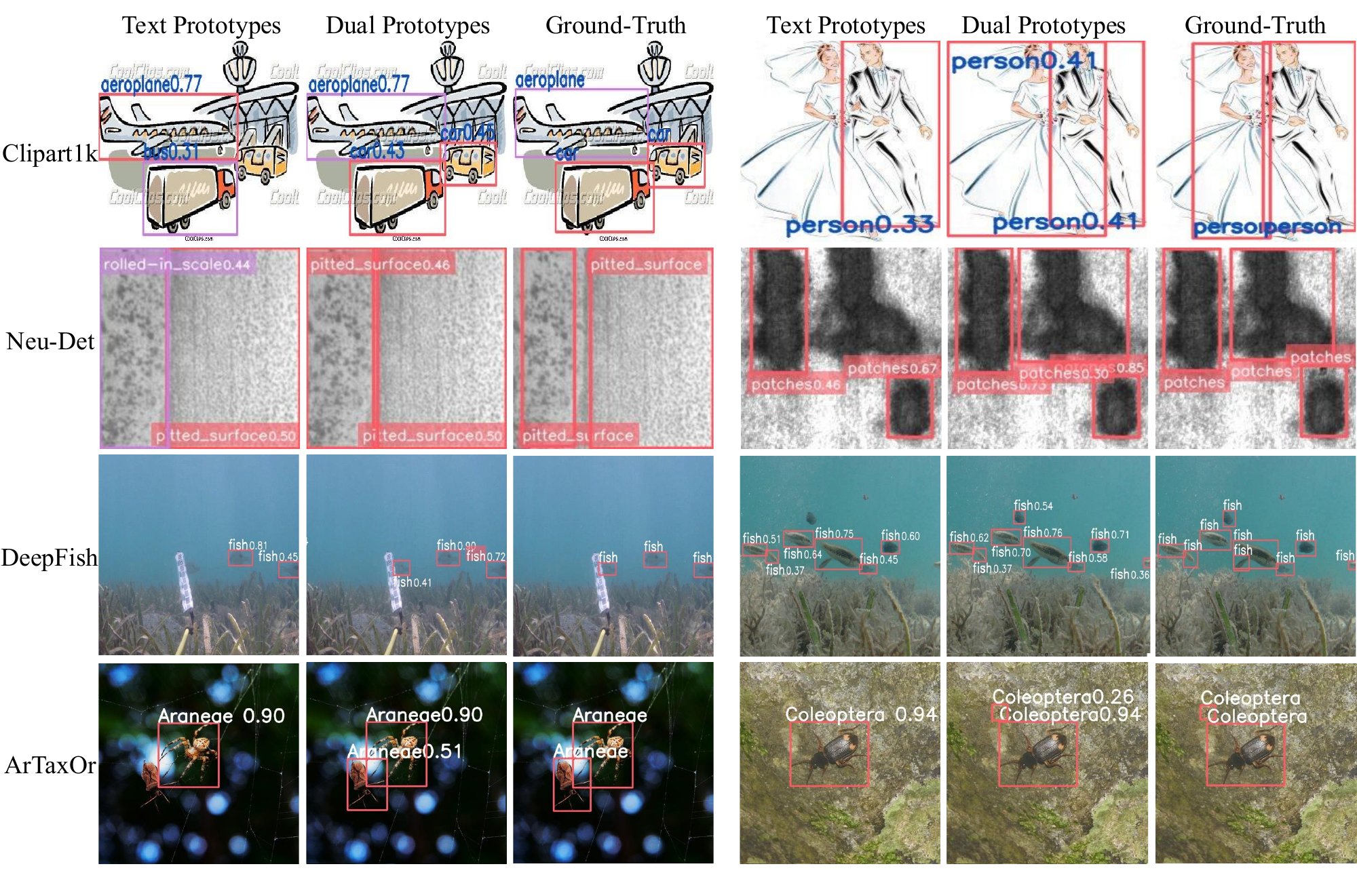}
  \caption{Qualitative comparison on four target domains. Each triplet shows detections from a text-only prototype baseline (left), our dual-branch method (middle), and ground truth (right). Our visual+text prototypes yield tighter boxes and fewer confusions: clipart scenes reduce spurious boxes on background objects; steel-surface images better separate fine-grained defects; underwater scenes recover more small fish; insect images avoid duplicate boxes and improve localization.}
  \label{fig:bbox}
\end{figure*}

\vspace{5pt}
\noindent\textbf{Prototype-Space Visualization.}
We project the visual features from the visual branch to 2D space with t-SNE. As shown in Figure \ref{fig:tsne}, the triangles denote query features of matched detections and circles denote hard negatives. Different colors indicate class labels.
Unlike closed-set classification, these classes do not collapse into clusters. This is expected for CD-FSOD, as the model is optimized for detection with few shots and domain shift, so global manifold separation is not explicitly enforced. Also, prototypes guide local decisions rather than enforcing a global margin. 
Even without clear clusters, we observe that, first, many query features lie nearer to their corresponding class neighborhoods than to unrelated ones, consistent with prototype-aligned scoring in the decoder. Second, hard negatives interleave along mixed regions where categories are visually confusable precisely where false positives arise, which illustrates why explicit negative prototypes help the detector suppress spurious responses.


\vspace{5pt}
\noindent\textbf{Qualitative Analysis of Prototype Guidance.}
Figure \ref{fig:bbox} visualizes how visual prototypes and hard-negative prototypes change the detector’s behaviour compared with a text-only baseline. 
In Clipart1k, text-only prompts often focus on context, e.g., terminal structures or buses. Visual prototypes down-weight these distractors, which leaves only the correct objects.
In NEU-Det, which contains industrial textures, the baseline confuses visually similar categories, but our method sharpens boundaries and assigns correct labels.  In underwater imagery dataset DeepFish, which contains small, low-contrast fish. The prototype guidance increases recall and reduces missed detections.
For insects in ArTaxOr, our method produces tighter boxes and avoids missing detections. 

%% file: sec/5_Conclusion.tex
\section{Conclusion and Limitations}
In this paper, we presented a dual-branch framework for cross-domain few-shot object detection that learns multi-modal prototypes. The text branch preserves open-vocabulary semantics, while the visual branch injects domain-conditioned visual information through a Visual Prototype Construction module that unifies class-level prototypes with query-aware hard negatives. Built on a feature enhancer, similarity-based query selection, and a prototype-aware decoder, our design performs explicit hard-negative mining with standard focal loss, where no extra contrastive terms are required. Across diverse target domains and 1/5/10-shot settings, the approach delivers consistent gains, with the largest improvements in the extreme 1-shot regime. Ablations confirm the benefit of both components, including visual prototypes and hard negatives, and of initializing the visual branch from the text branch.

The limitations of this work include sensitivity to non-typical supports and the overhead from running two branches. Future work will explore adaptive prototype creation and pruning, stronger negative mining, e.g., ring/context regions and proposal-similarity distractors, lightweight support augmentation, and distillation to a single deployment branch. Extending the idea to richer textual prompts, semi-supervised support sets, and video or multi-view settings are further promising directions. 


%% file: sec/X_suppl.tex
\clearpage
\setcounter{page}{1}
\maketitlesupplementary
\appendix

\section{More Implementation Details}
\noindent\textbf{Loss and Matching Configuration.}
We adopt Hungarian matching with a combination of classification, L1 box regression, and GIoU costs, weighted at 1.0, 5.0, and 2.0, respectively. After matching, the final detection loss applies weights of 2.0, 5.0, and 2.0 for the same components. These configurations are used consistently for both the text-guided and visual-guided branches, ensuring balanced optimization between classification reliability and spatial localization.

\noindent\textbf{Training Augmentation Strategy.}
During training, we follow the data augmentation strategies used in GroundingDINO~\cite{liu2024grounding}. Specifically, each image undergoes random horizontal flipping, then follows one of two paths: it is either resized to a randomly sampled scale with size constraints, or it passes through a stronger augmentation sequence involving initial resizing, random-size cropping, and a final multi-scale resize. These augmentations improve the robustness of both the text-guided and visual-guided branches under few-shot supervision. During inference, all augmentations are disabled, and evaluation strictly follows COCO-style metrics.

\section{More Results}
\noindent\textbf{Results on Remote Sensing FSOD.}
We further evaluate our method on the NWPU VHR-10 remote sensing dataset~\cite{cheng2014multi}, a high-resolution optical remote sensing benchmark with 10 object categories. Following the class split in~\cite{liu2024few},  we treat \emph{airplane}, \emph{baseball diamond}, and \emph{tennis court} as novel classes, while the remaining 7 classes serve 
as base classes.  Experiments are conducted under the 3/5/10/20-shot settings.

Table~\ref{tab:rs-fsod} summarizes the results on NWPU VHR-10 following the evaluation setup of Domain-RAG~\cite{li2025domain}. The table is organized into two settings: (i) \emph{Upper part (standard RS-FSOD protocol)}: models are first trained on the 7 base classes with sufficient annotations, and then fine-tuned on the 3 novel classes using K-shot samples. In this setting, Domain-RAG applies its augmentation strategy on top of SAE-FSDet~\cite{liu2024few}; (ii) \emph{Lower part (CD-FSOD protocol)}: following the open-source cross-domain setting, the pretrained model is directly fine-tuned on all 10 classes simultaneously, where both base and novel classes contain only K labeled samples per class. For fair comparison with the first setting, mAP is computed exclusively on the three novel categories in both protocols.

Across all shot settings, our method achieves competitive performance and demonstrates generalization on remote sensing images, where objects are small with limited texture and surrounded by visually confusable backgrounds.

\begin{table*}[t]
\centering
\setlength{\tabcolsep}{6pt}
\caption{Main results (mAP) on  NWPU VHR-10 benchmark under the 3/5/10/20-shot settings. The upper part follows the standard RS-FSOD protocol, while the lower part follows the cross-domain setting.}
\label{tab:rs-fsod}
\begin{tabular}{l l l c c c c c}
\toprule
Method & Training Setting & Backbone & 3-shot & 5-shot & 10-shot & 20-shot & Average \\
\midrule
Meta-RCNN~\cite{yan2019meta}           & RS-FSOD & ResNet-50 & 20.51 & 21.77 & 26.98 & 28.24 & 24.38 \\
FsDetView~\cite{kang2019few}           & RS-FSOD & ResNet-50 & 24.56 & 29.55 & 31.77 & 32.73 & 29.65 \\
TFA w/cos~\cite{wang2020frustratingly}            & RS-FSOD & ResNet-50 & 16.17 & 20.49 & 21.22 & 21.57 & 19.86 \\
P-CNN~\cite{cheng2021prototype}       & RS-FSOD & ResNet-50 & 41.80 & 49.17 & 63.29 & 66.83 & 55.27 \\
FSOD~\cite{fan2020few}                 & RS-FSOD & ResNet-50 & 10.95 & 15.13 & 16.23 & 17.11 & 14.86 \\
FSCE~\cite{sun2021fsce}                 & RS-FSOD & ResNet-50 & 41.63 & 48.80 & 59.97 & 79.60 & 57.50 \\
ICPE~\cite{lu2023breaking}              & RS-FSOD & ResNet-50 &  6.10 &  9.10 & 12.00 & 12.20 &  9.85 \\
VFA~\cite{han2023few}                  & RS-FSOD & ResNet-50 & 13.14 & 15.08 & 13.89 & 20.18 & 15.57 \\
SAE-FSDet~\cite{liu2024few}           & RS-FSOD & ResNet-50 & 57.96 & 59.40 & 71.02 &\textbf{85.08} & 68.36 \\
Domain-RAG~\cite{li2025domain}          & RS-FSOD & ResNet-50 &\textbf{ 59.99} & \textbf{65.78} & \textbf{72.87} & 84.05 & \textbf{70.67} \\
\midrule
GroundingDINO~\cite{liu2024grounding}      & CD-FSOD & Swin-B    & 57.1  & 61.3  & 65.1  & 69.5  & 63.3 \\
Domain-RAG~\cite{li2025domain}          & CD-FSOD & Swin-B    & 58.2  & 62.1  & \textbf{66.6}  & 69.7  & 64.2 \\
\textbf{LMP (ours)} & CD-FSOD & Swin-B    & \textbf{63.9}  &\textbf{ 65.3}  & 66.4  & \textbf{70.4}  & \textbf{66.5}  \\
\bottomrule
\end{tabular}
\vspace{-1pt}
\end{table*}

\noindent\textbf{Ablation on Multi-Scale Feature Levels.} 
GroundingDINO's Swin-B backbone follows the standard DETR practice and outputs a three-level feature pyramid at 1/8, 1/16, and 1/32 of the input resolution, corresponding to the last three Swin stages. GroundingDINO constructs a fourth level at 1/64 via downsampling to enrich the multi-scale representation. We denote these four levels as Level-0 to Level-3 in order of decreasing resolution. Level-0 corresponds to the ``lowest pyramid level'' referenced in the main paper, where we extract RoI features from the finest-resolution backbone output.

Table~\ref{tab:pyramid} reports 5-shot results across six target domains using different feature levels for visual prototype construction. All feature levels outperform the text-only baseline (40.8 mAP), confirming that domain-conditioned visual information complements semantic text features. On average, Level-0 achieves the best performance (44.0 mAP), as its fine-grained spatial resolution captures precise object boundaries and texture patterns critical for localization under domain shift. 

However, optimal feature levels vary across datasets depending on their visual characteristics. Clipart1k achieves comparable performance across levels, with Level-3 reaching 60.7 mAP and Level-0 reaching 60.1 mAP, as cartoon images with stylized shapes can benefit from both semantic abstraction and fine-grained details. Conversely, NEU-DET and UODD show peak performance at Level-1 (26.0 and 28.9 mAP), as industrial defects and underwater objects with low contrast and ambiguous boundaries require intermediate-level features balancing spatial precision and contextual understanding. These observations suggest that multi-level feature aggregation could further improve prototype construction by adapting to diverse visual characteristics, which we leave for future work.

\begin{figure*}[!t] 
  \centering
  \includegraphics[width=0.88\linewidth]{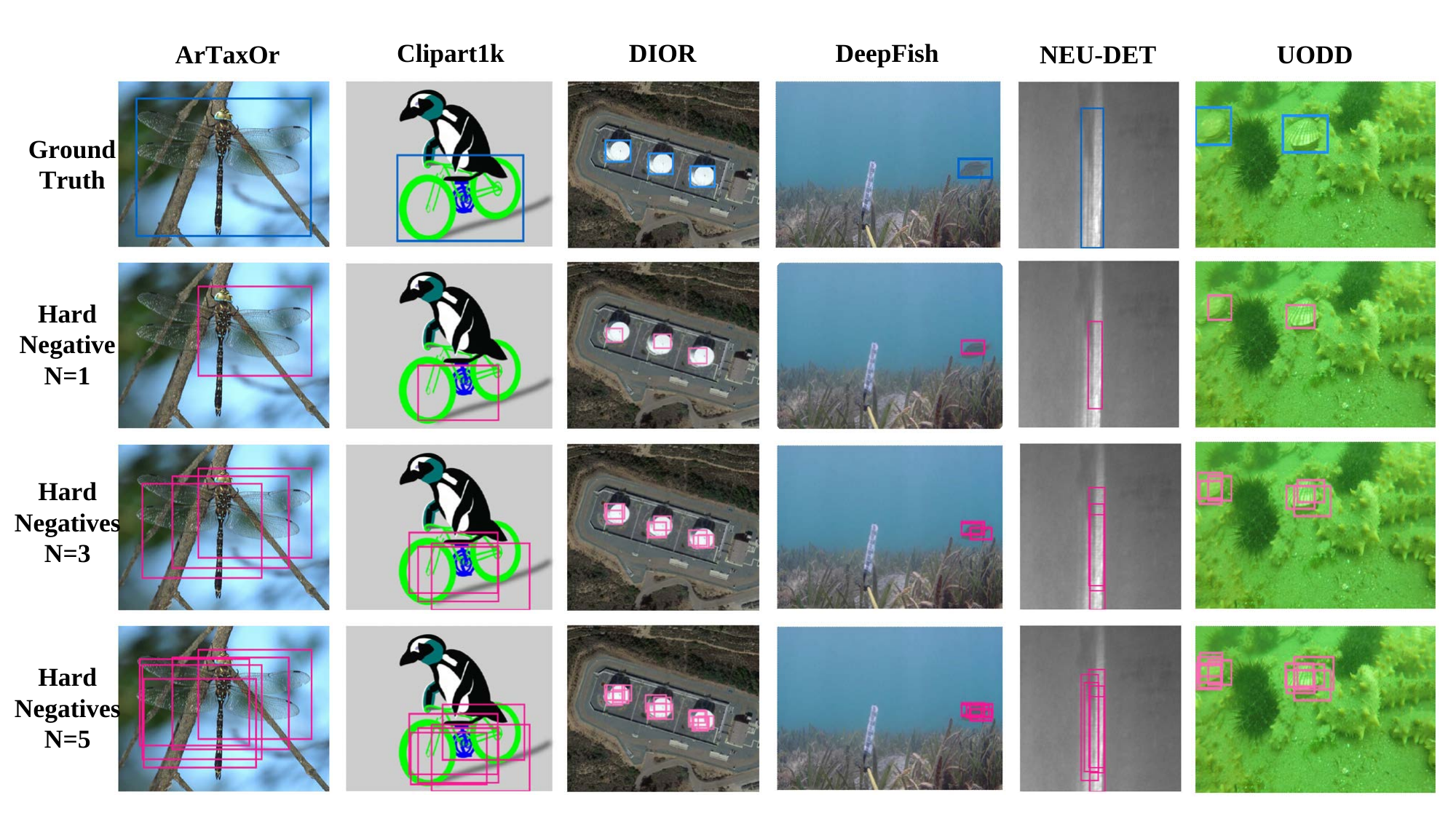}
   \caption{Visualization of hard negative sampling across six target domains. From top to bottom: ground truth and corresponding hard negatives with $N=1/3/5$ ($\operatorname{IoU}\!\in\![0.1,0.5]$ with GT).  With 
 $N=1$,  only a single boundary region is covered. $N=3$ achieves complementary spatial coverage across diverse object parts. $N=5$ produces redundant negatives that collapse into overlapping areas.}
  \label{fig:har-negative-bbox}
\end{figure*}

\noindent\textbf{Hard Negative Visualization.} Figure~\ref{fig:har-negative-bbox} illustrates how hard negatives capture partial object regions across target domains. By jittering ground-truth boxes with $\operatorname{IoU}\!\in\![0.1,0.5]$, each sampled region retains incomplete object evidence concentrated around confusable boundaries where false positives frequently arise. When $N=1$, only a single object part (e.g., head or tail) is covered, leaving other ambiguous regions unmodeled. With $N=3$, the jittered boxes span multiple object parts and boundary variations, providing a compact yet comprehensive coverage that explains the performance peak observed in Figure~\ref{fig:hyper-parameter} of the main paper. Increasing beyond $N=3$ produces redundant negatives that  duplicate boundary cues, diluting positive signals and ultimately degrading detection performance.
\begin{table*}[t]
\centering
\setlength{\tabcolsep}{6pt}
\caption{Ablation on multi-scale feature levels for visual prototype construction (5-shot). Baseline uses text-only prototypes. Level-0/1/2 use backbone outputs at 1/8, 1/16, 1/32 resolutions, and Level-3 uses downsampling to 1/64. Level-0   performs best on average, indicating that higher-resolution features preserve fine-grained appearance cues essential for discriminative visual prototypes under domain shift.}
\label{tab:pyramid}
\begin{tabular}{l c c c c c c c}
\toprule
Method    & ArTaxOr & Clipart1k & DIOR & DeepFish & NEU-DET & UODD &Average\\
\midrule
Baseline  & 67.8    & 58.3    & 29.5   & 42.1   & 20.7   & 26.5 &40.8\\
Level-0 (ours)
& 75.0    & 60.1    &\textbf{ 31.3}   & \textbf{43.9}   & 25.1  & 28.3 &\textbf{44.0}\\
Level-1 
& 70.5    & 59.8    & 30.4   & 43.3   &\textbf{ 26.0}  & \textbf{28.9}&43.2\\
Level-2 
& \textbf{75.9}    & 59.2    & 29.6   & 42.8   & 21.4  & 27.8 &42.8\\
Level-3  
& 73.8    &\textbf{ 60.7}    & 29.8   & 42.5   & 23.5  & 27.2 &42.9\\
\bottomrule
\end{tabular}
\vspace{-5pt}  
\end{table*}

